\documentclass[a4paper, 10pt, conference]{ieeeconf}      
\usepackage{FG2024}
\usepackage{multirow} 
\usepackage{booktabs}
\usepackage{graphicx}

\IEEEoverridecommandlockouts                              
\def\FGPaperID{****} 

\title{\LARGE \bf
Real, fake and synthetic faces - does the coin have three sides?
}

\author{\parbox{16cm}{\centering
    {\large Huibert Kwakernaak$^1$ and Pradeep Misra$^2$}\\
    {\normalsize
    $^1$ Faculty of Electrical Engineering, Mathematics and Computer Science, University of Twente, Enschede, The Netherlands\\
    $^2$ Department of Electrical Engineering, Wright State University, Dayton, USA}}
    \thanks{This work was not supported by any organization}
}

\begin{document}

\ifFGfinal
\thispagestyle{empty}
\pagestyle{empty}
\else
\author{\parbox{16cm}{\centering
    {\large Shahzeb Naeem$^1$$^3$, Ramzi Al-Sharawi$^1$, Muhammad Riyyan Khan$^1$, Usman Tariq$^1$$^4$, Abhinav Dhall$^2$ and Hasan Al-Nashash$^1$}\\
    {\normalsize
    $^1$American University of Sharjah, University City, Sharjah and 26666, United Arab Emirates\\
    $^2$Flinders University,  Sturt Rd, Bedford Park South Australia and 5042, Australia}\\
    Email: {\normalsize
    $^3$b00080174@aus.edu, $^4$utariq@aus.edu}}
}
\pagestyle{plain}
\fi
\maketitle



\begin{abstract}

With the ever-growing power of generative artificial intelligence, deepfake and artificially generated (synthetic) media have continued to spread online, which creates various ethical and moral concerns regarding their usage. To tackle this, we thus present a novel exploration of the trends and patterns observed in real, deepfake and synthetic facial images. The proposed analysis is done in two parts: firstly, we incorporate eight deep learning models and analyze their performances in distinguishing between the three classes of images. Next, we look to further delve into the similarities and differences between these three sets of images by investigating their image properties both in the context of the entire image as well as in the context of specific regions within the image. ANOVA test was also performed and provided further clarity amongst the patterns associated between the images of the three classes. From our findings, we observe that the investigated deep-learning models found it easier to detect synthetic facial images, with the ViT Patch-16 model performing best on this task with a class-averaged sensitivity, specificity, precision, and accuracy of 97.37\%, 98.69\%, 97.48\%, and 98.25\%, respectively. This observation was supported by further analysis of various image properties. We saw noticeable differences across the three category of images. This analysis can help us build better algorithms for facial image generation, and also shows that synthetic, deepfake and real face images are indeed three different classes.

\end{abstract}

\section{INTRODUCTION}

The advent of readily available deep learning models, e.g. Generative Adversarial Networks (GANs) \cite{1}, have simplified and popularized the generation of deepfake media, with various works within literature looking to build upon the findings of their predecessors to further enhance the verisimilitude of deepfakes. Deepfake facial manipulation techniques can be classified under identity swapping, attribute manipulation, and expression swapping \cite{2}. The simplest approach, which has particularly been followed by early deepfake algorithms, typically involves making use of two encoder-decoder pairs: each encoder is trained firstly on the source image, then the target image. Once training is complete, the decoders of the source image and target image are swapped such that the encoder of the source image is paired with the decoder of the target image \cite{3}, which generates the source image on top of the target image while retaining the target image’s features. This approach, though simple, is used as baseline for subsequent deepfake generation algorithms in the literature. The deepfake generation has continued to grow and produce increasingly realistic deepfake content, e.g. generation of deepfakes with GANs \cite{seow2022comprehensive} and diffusion models \cite{Ivanovska_2024_WACV}. \par

With recent advancements in generative AI, content synthesis, which refers to media that is fully generated by AI, has also emerged as an ever-evolving subfield of fake media generation and detection. This has culminated towards the flagship of such work in the form of OpenAI’s powerful AI model “Sora”, which creates realistic and imaginative scenes from a user-inputted text prompt \cite{videoworldsimulators2024}. There has been a lot of work on synthetic image generation, with various works seeking to generate synthesized images based on user-input text prompts \cite{4}. While showcasing the ever-growing power of generative AI, such synthesis models raise similar ethical concerns as deepfake generation models due to their uses in spreading fake news, propaganda, and other malicious usage of media. \par

In this work, we define real face images as the images of real persons; deepfake/fake face images as the images in which there is some manipulation on real images, e.g. identity transfer etc; and synthetic face images as those which are generated from image generation algorithms without any image prompt. The question that we ask in this work is as follows: are real, deepfake and synthetic media indeed three different classes? If they are, then perhaps we need to develop different strategies in countering them. \par

To address this research question, we propose a novel method to analyze the aforementioned three types of images in two stages. First, we train various deep learning models and test their abilities to classify real, deepfake, and synthetic face images. Second, we conduct an analysis across various image properties such as brightness, sharpness, luminosity, RGB mean, contrast, and detail for all three sets of images to further understand how some of the image property values change across real, deepfake, and synthetic face images. Previous works within literature have only sought to analyze the differences between real images and deepfake images \cite{KOSARKAR20232636}; therefore, the novelty of our analysis will help understand differences across the three classes in general, and between the deepfake and synthetic faces in particular. This information can then be used to develop resilient detection algorithms and can also improve the quality of generation algorithms. It is worth noting that, synthetic media generation is an older field in comparison to deepfakes generation. It has been widely used in the entertainment industry. This is in contrast to deepfakes, which have much more malicious usages. Hence, it makes sense to treat deepfakes and synthetic media as two different classes. In this work, we are looking for the experimental evidence for this difference, in addition to their differences in comparison to real face images. \par

The rest of this paper is structured as follows: Section II briefly discusses relevant works within literature regarding deepfake generation and detection and synthetic face image generation and detection. Section III then details the dataset used and the pre-processing techniques. Section IV outlines the methodology, while Section V presents results and discussion. Finally, Section VII ends this paper with concluding remarks and a brief discussion into future work on this topic. \par

\section{RELATED WORKS}

\subsection{Deepfake Generation and Detection}

Building upon the basic building blocks of deepfake generation algorithms, Choi et al. \cite{5} developed StarGAN, which looks to tackle the shortcomings of traditional basic deepfake algorithms in multi-domain image translation tasks. The StarGAN architecture involves training a single generator G to learn mappings among multiple domains \cite{5}. An auxiliary classifier is also introduced to allow the single discriminator in Choi et al.’s GAN architecture to control multiple domains \cite{5}.  Quantitative analysis on the results of their StarGAN architecture demonstrated that it outperformed DIAT \cite{6}, IcGAN \cite{7} and CycleGAN \cite{8} in single attribute transfer tasks, multi-attribute transfer tasks, and in terms of facial expression classification error. \par

Developing on the StarGAN architecture presented in \cite{5}. He et al. \cite{9} developed AttGAN, which only edits the facial attributes of an input image based on the desired effect. It does this by using an encoder-decoder architecture in conjunction with an adversarial learning component. Typically, facial attribute editing is based on an encoder-decoder architecture and is achieved by decoding the latent representation of the given face conditioned on the desired attributes. Some works have looked to establish an attribute-independent latent representation for further attribute editing, although these are prone to information loss \cite{9}. Consequently, He et al. apply attribute classification constraints and reconstruction learning to the generated image to only guarantee the change of the desired attributes. The adversarial learning component of their AttGAN architecture is then used for visually realistic editing. The authors of \cite{9} compare their AttGAN architecture with other notable deepfake architectures such as StarGAN \cite{5}, VAE/GAN \cite{10}, IcGAN \cite{7}, Fader Networks \cite{11}, CycleGAN \cite{8}, and the architecture proposed by Shen et al. \cite{12}. They showed that AttGAN yielded higher attribute editing accuracy and lower attribute preservation error.\par

To tackle the issue of malicious deepfake generation, researchers have looked towards deep learning and machine learning for inspiration into building deepfake detection models. For example, \cite{13} utilizes the pre-existing XceptionNet model for deepfake detection, obtaining an accuracy of 91.9\%, precision of 91.7\%, and recall of 92.2\% through training for 20 epochs on the Faceforensics++ dataset \cite{14}. Ilyas et al. \cite{15} also present a novel approach involving the fusion of two EfficentNet-b0 models, one with the ReLU activation function and the other with the Swish activation function. Their proposed architecture yielded high detection results with an accuracy and precision of 96.5\% and 97.07\%, respectively, when trained on the Faceforensics++ \cite{14} dataset and 88.41\% and 91.28\%, respectively, when trained on the DFDC-preview \cite{dolhansky_deepfake_2019} dataset. Furthermore, Guarnera et al. proposed detecting deepfakes by analyzing convolutional traces, which involved the development of an Expectation Maximization algorithm which outputs a feature vector representing the structure of the Transpose Convolutional Layers used in generating a deepfake image \cite{16}. This algorithm was used to encode generated images from traditional deepfake algorithms such as StarGAN \cite{5}, AttGAN \cite{9}, GDWCT \cite{17}, StyleGAN \cite{18}, and StyleGAN2 \cite{19}, before being passed into traditional machine learning algorithms. It was shown that Guarnera et al.’s algorithm performed best when using SVM to classify deepfake images generated by StyleGAN2 with a 4x4 kernel, which yielded a classification accuracy of 99.81\% \cite{16}. The results of such deep learning-based deepfake detection algorithms highlight their potential in accurately classifying deepfake content. \par

Other works have focused on more innovative approaches to this problem: one such approach was proposed by Cai et al. in the form of MARLIN, which is a masked autoencoder for facial video representation learning \cite{20}. The results of \cite{20} showcase MARLIN’s excellence as a facial video encoder and feature extractor, which allow it to be used in tasks like facial attribute recognition, facial expression recognition, deepfake detection, and lip synchronization, among others. To compare with other state-of-the-art (SOTA) methods, Cai et al. evaluated MARLIN's performance in deepfake detection, yielding an accuracy of 89.43\% and an area under the curve (AUC) score of 0.9305 on the Faceforensics++ dataset \cite{20}. 

\subsection{Synthetic Image Generation and Detection}

The development and use of synthetic generation techniques has emerged within literature for a plethora of applications, one of which is for facial animation generation. For example, text-guided media generation techniques have presented themselves in the forms of works such as that proposed by Xia et al. \cite{4}, Reed et al. \cite{26}, and Zhang et al. \cite{zhang2017stackgan}. Reed et al. \cite{26} propose a text-to-image synthesis model called GAN-INT-CLS and advocate for the efficacy of their model by showcasing its ability to produce synthesized images of birds and flowers from textual descriptions. Zhang et al. \cite{zhang2017stackgan} build upon this work to develop StackGAN for text-to-image synthesis; using the same methodology as that in \cite{26}, Zhang et al. perform a quantitative analysis to demonstrate StackGAN’s superiority over GAN-INT-CLS \cite{26} and GAWWN \cite{reed2016learning} in terms of inception scores and human rank. Whereas \cite{26} and \cite{zhang2017stackgan} focus on general text-to-image synthesis, Xia et al. \cite{4} propose a text-to-image synthesis model called TediGAN and use it for the purposes of facial image synthesis. From their quantitative analysis, \cite{4} showcases TediGAN’s superiority over other SOTA models such as AttnGAN \cite{xu2017attngan}, ControlGAN \cite{li2019controllable}, DFGAN \cite{tao2022dfgan}, and DM-GAN \cite{zhu2019dmgan}. Overall, all of the discussed works present promising results with regards to the development and application of text-to-image synthesis models.

A good body of literature has focused on image synthesis for data augmentation in medical applications. One such approach was developed by Waqas et al. \cite{22}, who proposed an enhanced variant of PGGAN \cite{23}, Enhanced-GAN, that includes a self-attention layer and spectral normalization to improve stability during training. The authors then show that EnhancedGAN yields better AM and Mode scores compared to PGGAN at generation of 128 and 256 resolution images \cite{22}, in addition to its ability to generate high-quality synthesized knee MR images. \par

Another approach by Liang et al. has sought to perform ultrasound image synthesis via sketch guided PGGANs \cite{24}. To do this, Liang et al. adopt a sketch GAN (SGAN) to introduce auxiliary sketch guidance upon object mask in a conditional GAN (cGAN), before implanting the SGAN into a PGGAN to yield their proposed PGSGAN algorithm that can gradually synthesize ultrasound images from low to high resolution. The authors of \cite{24} then compare the performance of their proposed PGSGAN algorithm with cGAN and SGAN using the Freshet Inception Distance (FID), Kernel Inception Distance (KID), and multi-scale structural similarity (MS-SSIM) as metrics. The results of their analysis indicate the superiority of their proposed PGSGAN algorithm compared to the other ones investigated in their study, with an FID 54.94, a KID of 412, and an MS-SSIM of 0.4895. \par

Although the aforementioned biomedical applications represent the good of image synthesis generation, text-to-image synthesis has also progressed within literature in generating realistic images and videos revolving around everyday life, as previously discussed. Such models have can have malicious applications, such as the spread of fake news, identity theft, and blackmail. This necessitates the development of detection models to detect such synthesized media. FakeSpotter, proposed by Wang et al. \cite{27}, is one of such models, which detects AI-synthesized fake faces by capturing the layer-wise activated neurons in their deep neural network with a novel neuron coverage criterion termed neuron mean coverage (NMC). By testing their model on generated images from SOTA generative models such as StyleGAN \cite{18}, StyleGAN2 \cite{19}, and PGGAN \cite{23}, Wang et al. showcased their model’s superiority in detecting AI-synthesized fake faces, with a classification accuracy of 90.6\%, thus showcasing its ability for detecting AI-synthesized faces. AI-synthesis detection is a recent area, with various works within literature have begun looking into this problem \cite{28}. \par

\section{Dataset and Data Preprocessing}

To address this papers’s goals, we used three categories of images for model training and subsequent analysis: real face images, deepfake face images, and synthetic face images. A subset of publicly available SFHQ-1 dataset \cite{29}, which contains 89,785 high-quality synthesized 1024x1024 curated face images, was used for model training and analysis related to synthetic images. The “140k Real and Fake Faces” dataset from Kaggle \cite{30} was used for model training and analysis related to deepfake and real images. This dataset contains 70,000 real images sampled from the Flickr dataset generated by NVIDIA \cite{18}, in addition to 70,000 deepfake images sampled from Bojan Tunguz’s “1 Million Fake Faces” dataset \cite{31}. The images were then passed through a preprocessing stage prior to training the deep learning models. \par

The images within the dataset had to be preprocessed to ensure that the pose and facial crop styles were consistent throughout the dataset. This was done to mitigate the information coming e.g. from background or pose variations across the datasets, that could help the deep learning algorithms. We used the YOLOv8 Nano (YOLOv8n) \cite{32} (pre-trained on the COCO dataset \cite{33}), for face detection, and YOLOv8n-pose \cite{32} (pre-trained on the COCO-Pose dataset \cite{33}) for pose identification. We constrained the images in our dataset to be frontal or near frontal.  \par

Following this, the distance from the left eye to the nose, in addition to the distance from the right eye to the nose, was calculated; ideally, when a face is in the frontal pose, the ratio between these distances will approximately one. Therefore, by setting a lower threshold of 0.9 and an upper threshold of 1.1, the frontal posed faces were extracted from all images across the real dataset, deepfake dataset, and synthetic dataset. We saved the extracted images as 256x256, to ensure consistency with regards to image dimensions within the dataset. Random extraction from all three datasets was then utilized to obtain 10,000 images per class for the training set, 2,000 images per class for the validation set, and 10,000 images per class for the testing set. \par

\section{Methodology}

\subsection{Deep Learning Models}

Image feature extraction is a pivotal task in investigating the patterns associated with real images, deepfake images, and synthetic images. We implemented eight well-known models within literature and analyzed their performances on our dataset. These models, arranged in order of complexity from most floating point operations per second (FLOPS) to least, are:  ViT Patch-16 \cite{39}, DenseNet-121 \cite{40}, ResNet-50 \cite{38} InceptionNet-v3 \cite{34}, EfficientNet-b0 \cite{37}, VGG-16 \cite{41}, ShuffleNet-v2 \cite{36}, and MobileNetV2 \cite{42}. All of these models were pre-trained on the ImageNet dataset \cite{35}.

Prior to training, various augmentations were applied to the images in the training set to prevent overfitting; these augmentation techniques included flipping, minor rotation, affine transformations, grayscale transformations, cropping, and color jitter. Furthermore, each image was resized to 224x224 before passing into each of the aforementioned models for training with the exception of InceptionNet-v3, given that it takes in 299x299 images as input.

\subsection{Image Property Analysis}

\begin{figure}[!t]
  \centering
  \includegraphics[scale=.4]{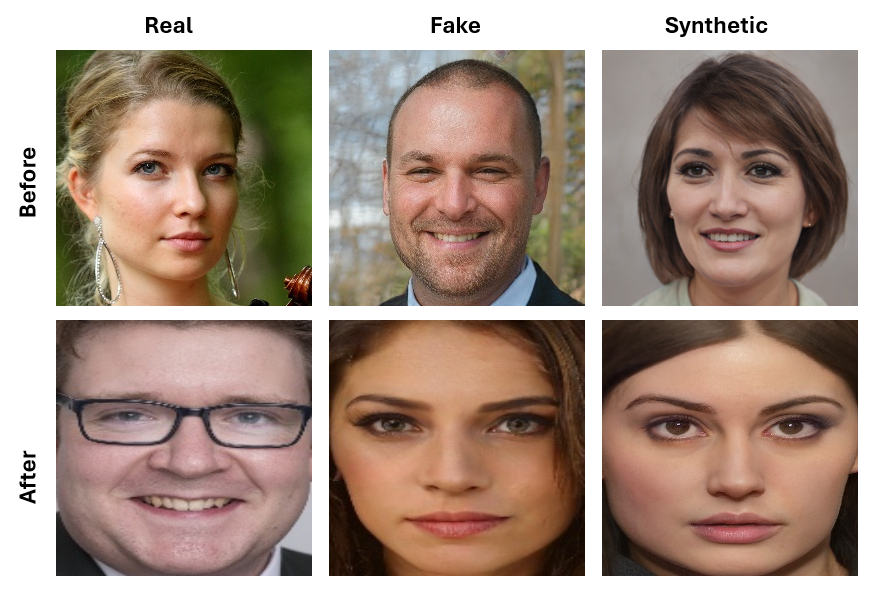}
  \caption{Sample images before and after preprocessing to focus on facial features}
  \label{fig:1}
\end{figure}

\begin{figure}[!t]
  \centering
  \includegraphics[scale=.25]{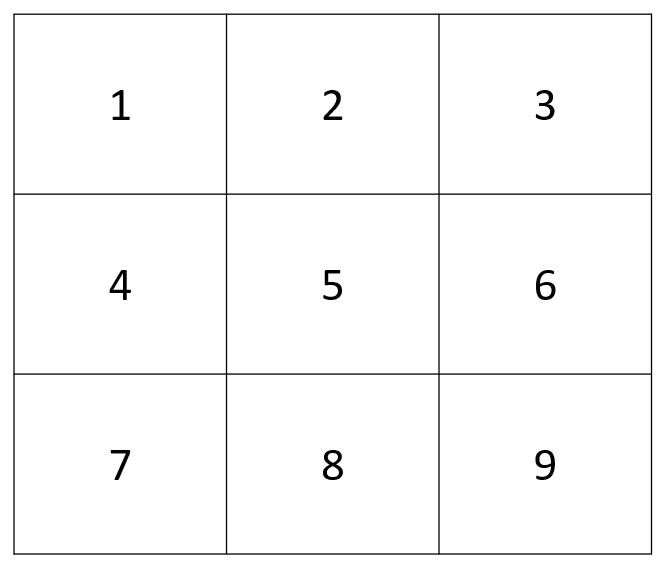}
  \caption{Illustration of regions in an image, corresponding to different regions of a subject's face}
  \label{fig:2}
\end{figure}

After investigating the performance of the deep learning models in distinguishing between the three different types of classes, we further seek to analyze the differences between real images, deepfake images, and synthetic images by analyzing various image properties. To do this, we first apply preprocessing techniques to focus on the subject's facial area; then, we divide each facial image into nine regions to break it down further, as shown in Fig. \ref{fig:1}. Analysis is performed on these non-overlapping portions of the image and on the entire image. Examples of real, deepfake, and synthetic image before and after such preprocessing is showcased in Fig. \ref{fig:2}.

The properties to be analyzed for each image and its nine regions are listed below (please note that “region” and “image” are used interchangeably):

\begin{enumerate}

    \item \textbf{Brightness}: this is computed by taking the mean pixel intensity of the grayscale image. It tells us how bright the overall image appears to human perception. The higher the mean intensity, the brighter the image and vice versa.
    
    \item\textbf{Sharpness}: this is determined using the Laplacian operator applied to the grayscale image. It calculates the second derivative of the image intensity, emphasizing regions with rapid intensity changes. The higher the Laplacian value, the more frequent the presence of sharp edges and transitions between different regions in the image.
    
    \item \textbf{Luminosity}: this is calculated from the L channel of the LAB color space, which separates color information into three channels: L (luminosity), A (green-red), and B (blue-yellow). The luminosity channel represents the perceived lightness of the image and is also applied to a grayscale image. The more the luminosity, the more the perceived lightness or intensity of the image.
    
    \item \textbf{RGB Mean}: this calculates the average intensity of each color channel (red, green, and blue) in the image. It provides insight into the dominant colors present in the image and their overall intensity levels.
    
    \item \textbf{Contrast}: this is computed as the standard deviation of pixel intensities in the grayscale image. It measures the variation in intensity levels across different regions of the image. The higher the contrast values, the greater the difference between light and dark areas in the image.
    
    \item \textbf{Detail}: this is derived from the Sobel operator, which performs edge detection by calculating the gradient magnitude, and reflects the level of texture and fine features present in the image. It emphasizes regions with significant changes in intensity, such as edges and contours. The higher the detail, the larger the presence of textures and fine features in an image.
    
\end{enumerate}

The patterns across the regions and images of the different classes, based off on the above properties, will be explored by plotting variations of each of these properties between classes as line plots. Furthermore, bar plots will be incorporated to visualize p-values for ANOVA. This will allow for the observation of various trends in the p-values across the different regions and entire images, leading to important insights as to whether the differences in various measures are due to random variations or due to significant differences across the three image classes.

\section{Results}

As outlined in the methodology, the analysis of trends between real images, deepfake images, and synthetic images was done in two parts. Firstly, the performance of traditional deep learning models on distinguishing between these three sets of images was investigated. Secondly, an analysis of image properties between real images, deepfake images, and synthetic images was conducted to investigate the trends associated between them in such properties. The results of these procedures are outlined below. \par

\subsection{Deep Learning Models}

\begin{figure}[!t]
  \centering
  \includegraphics[scale=.17]{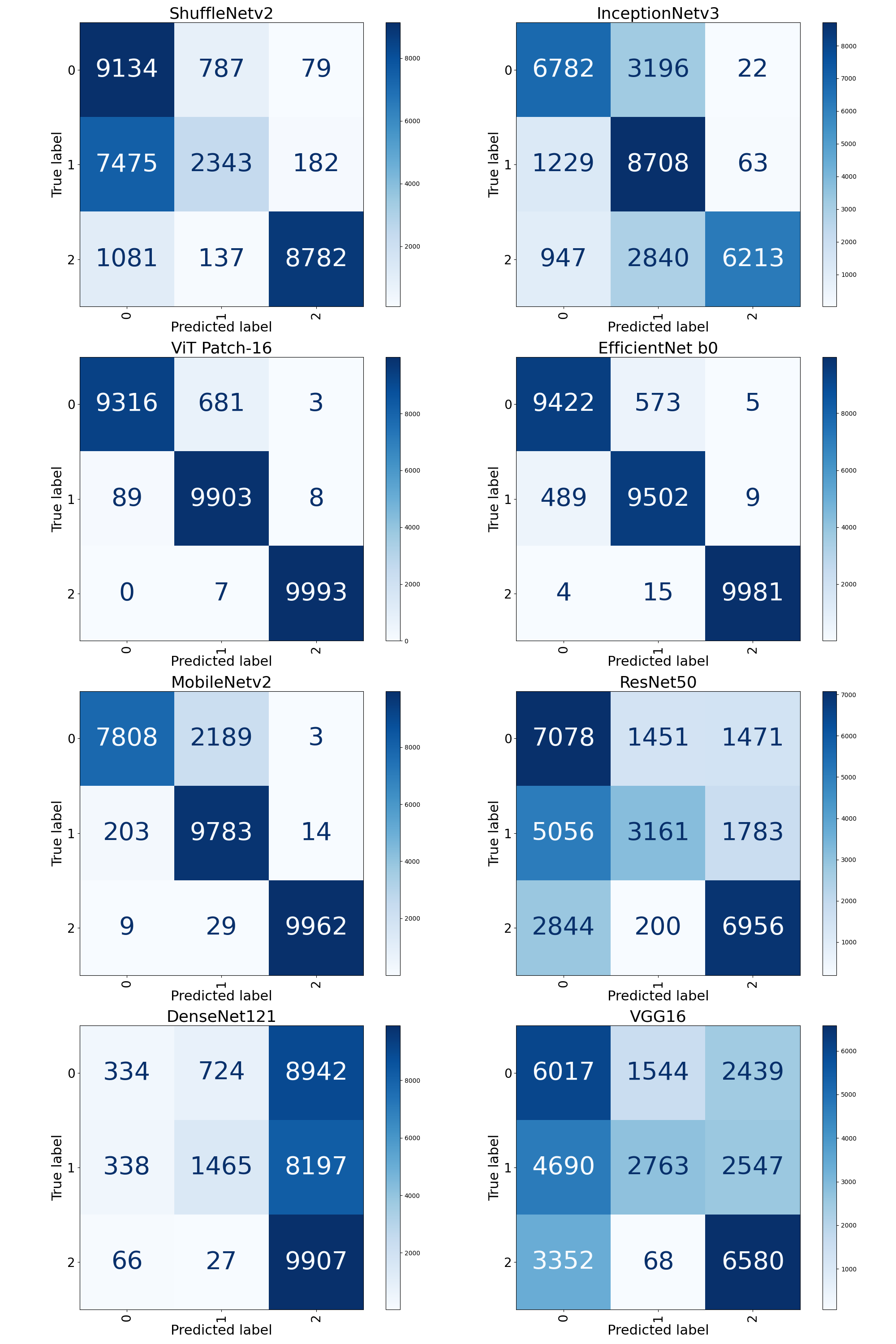}
  \caption{Confusion matrices for each model 
  (0 = deepfake, 1 = real, 2 = synthetic)}
  \label{fig:3}
\end{figure}

\begin{figure}[!t]
  \centering
  \includegraphics[scale=.11]{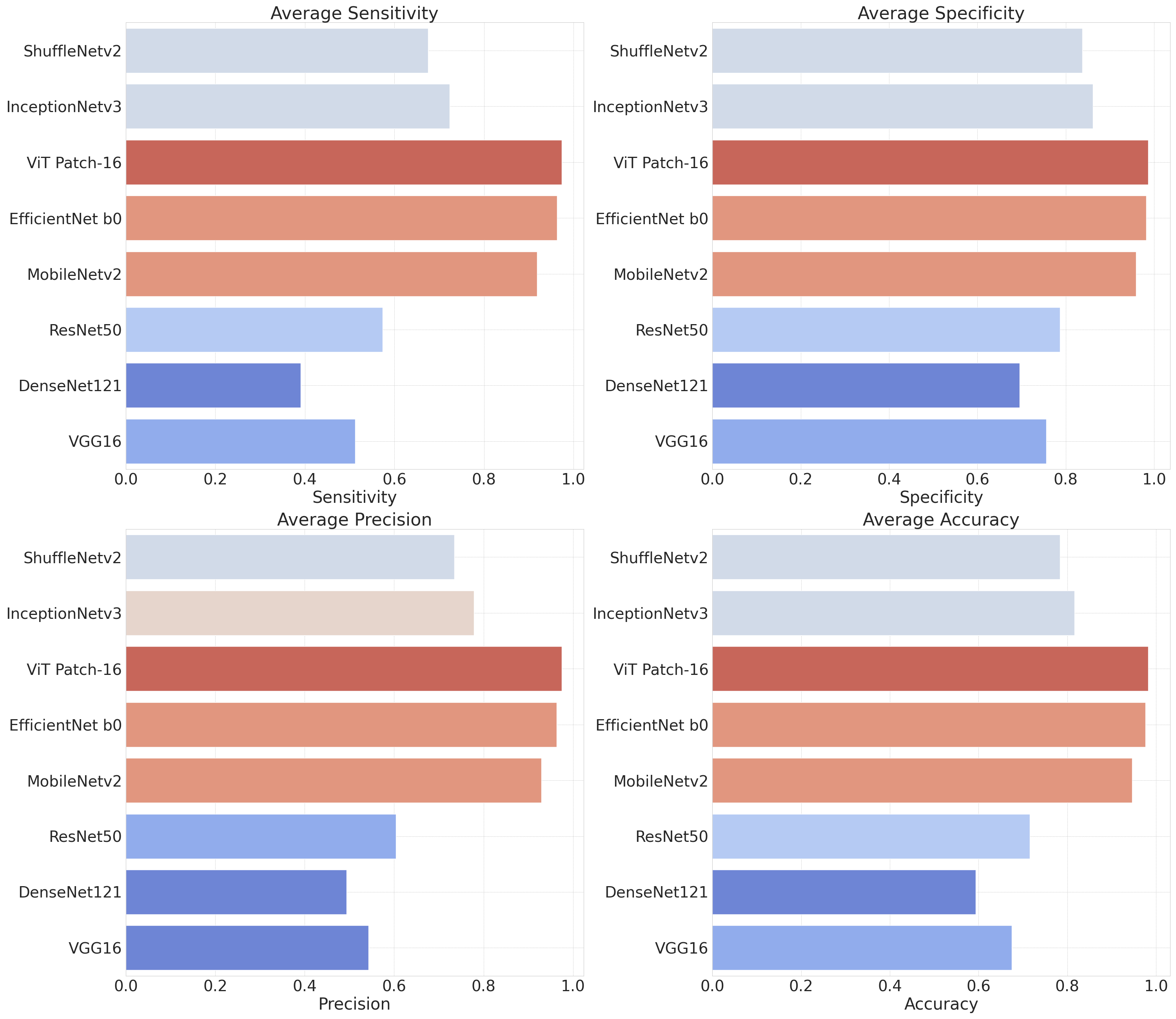}
  \caption{Class-averaged sensitivity, specificity, precision, and accuracy for each model}
  \label{fig:4}
\end{figure}

\begin{figure}[!t]
  \centering
  \includegraphics[scale=.17]{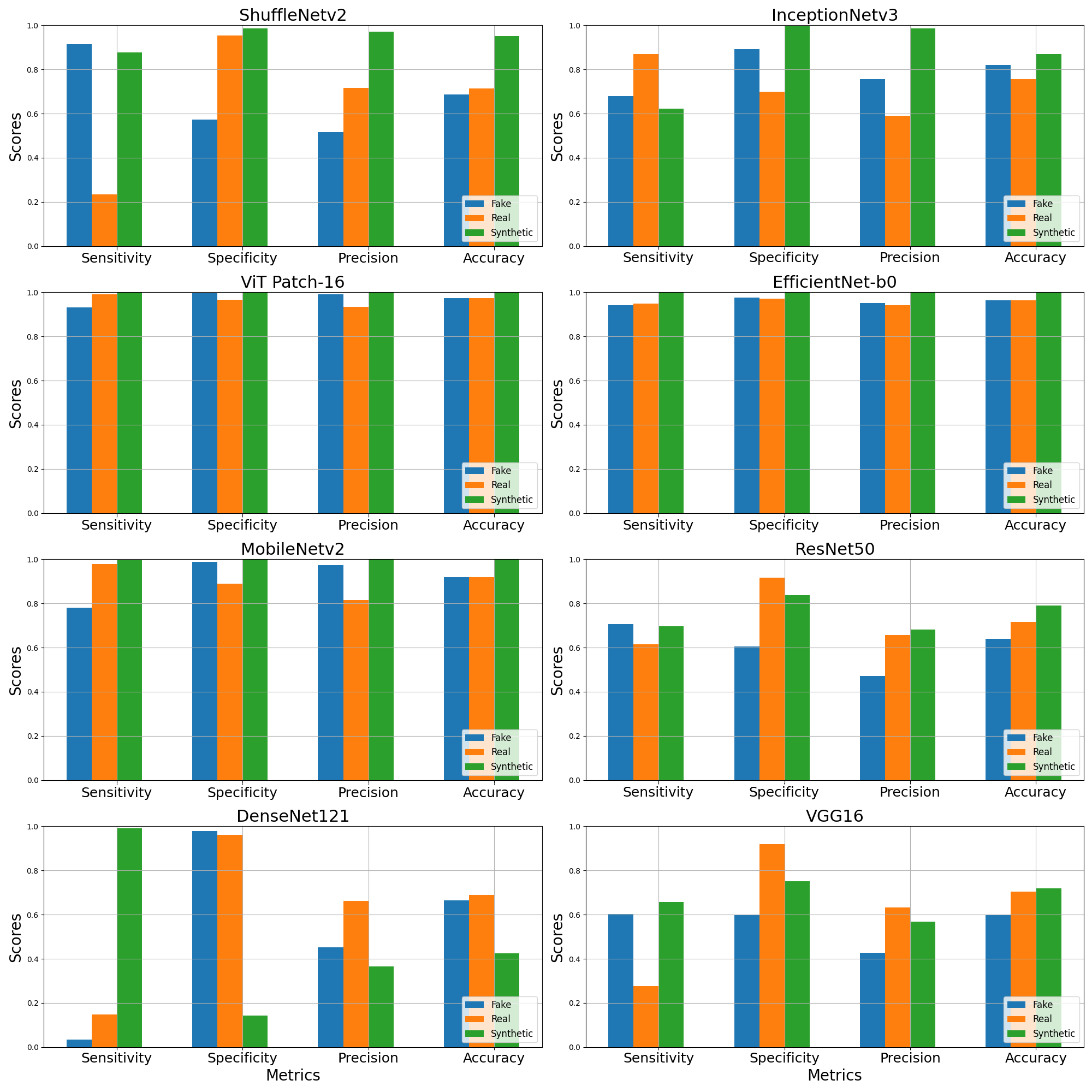}
  \caption{Sensitivity, specificity, precision, and accuracy for each class within each model}
  \label{fig:5}
\end{figure}

The classification results of the previously discussed eight deep learning models when tested on the dataset described earlier are outlined in Fig. \ref{fig:3}, Fig. \ref{fig:4}, and Fig. \ref{fig:5}. From these figures, it can be deduced that ViT Patch-16, EfficientNet-b0 and MobileNetv2 performed best on classifying between the three images classes. Amongst these models, Patch-16 excels performance-wise, with an average sensitivity of 97.37\%, an average specificity of 98.69\%, an average precision of 97.48\%, and an average testing accuracy of 98.25\%. 
The models that achieved good classification performance tended to be more complex compared to those that achieved bad classification performance, indicating the need for complex deep learning models to correctly classify between real images, deepfake images, and synthetic images. The average classification results for each of the eight models are summarized in Table \ref{tab:3}. \par

Another takeaway from Fig. \ref{fig:3} is that most of the tested models found it easier to detect synthetic images. This is backed up by prevailing higher values of sensitivity for the synthetic class in Table \ref{tab:1}, with values ranging from 62.13\% to 99.93\% across all models, as well as accuracy in Table \ref{tab:2}, with values ranging from 71.98\% to 99.94\% except for DenseNet-121, which has an accuracy of 42.56\% on the synthetic images. Furthermore, as shown in Fig \ref{fig:5} and Table \ref{tab:2}, we note that the precision for the synthetic class is typically the highest amongst all classes, or sometimes just slightly lower than that of the real class. This trend is not seen in DenseNet-121’s results, which can be attributed to being an anomaly as opposed to a norm. Overall, DensetNet-121 achieves the worst performance, as outlined in Fig. \ref{fig:4}, Fig. \ref{fig:5}, and Table \ref{tab:3}. It has the lowest accuracy, sensitivity, specificity, and precision as seen in Table \ref{tab:3}. In fact, the confusion matrix corresponding to DenseNet-121, shown in Fig. \ref{fig:3}, outlines that the model is simply predicting most of the images to be synthetic images; such a conclusion is supported by the poor average classification metrics for the DenseNet-121 model as shown in Table \ref{tab:3}. \par

When it comes to distinguishing between the different sets of images, it was shown that the trained models varied with regards to their performance in classifying between the different images, as best shown from the confusion matrices in Fig. \ref{fig:3}. For example, it can be seen that the DenseNet-121 model struggled in classifying real and deepfake images, and was instead biased towards classifying images as synthetic, which is further supported by the model's high recall for the synthetic class in Table \ref{tab:1}. Furthermore, the ShuffleNet-v2, VGG-16, and ResNet-50 models struggled in predicting images to be real as opposed to deepfake or synthetic, which is further supported by the low recall value for the real class compared to the other classes across all three models, as outlined in Table \ref{tab:1}. Finally, the InceptionNet-v3 model was found to be biased towards classifying images to be real as opposed to deepfake or synthetic, as further supported by the high recall value for the real class compared to the other classes as shown in Table \ref{tab:1}. Overall, it can be deduced from the results in Table \ref{tab:1} and Table \ref{tab:2} that the best-performing models were ViT Patch-16 and EfficientNet-b0, which were able to classify instances from all three classes with a high degree of accuracy. \par  


In summary, based on the presented results, we can conclude that the models found it easiest to detect synthetic facial images as opposed to images from the other two classes. However, some synthetic images might contain a high-level of verisimilitude, which can confuse some deep learning models. Furthermore, it was shown that the deep learning models were able to detect real images and deepfake images with similar levels of accuracy, which indicates the high level of similarity between the deepfake and real images. Both these takeaways are testament to the ever-rising quality of generative artificial intelligence, which necessitates continuous research into more powerful classification models that detect real images from deepfake and synthetic images. \par

\begin{table}[h]
\caption{Sensitivity and specificity metrics for each class across all trained models}
\label{tab:sensitivity-specificity}
\centering
\begin{tabular}{|l|l|c|c|}
\hline
\textbf{Model} & \textbf{Class} & \textbf{Sensitivity (Recall)} & \textbf{Specificity} \\ \hline
\multirow{4}{*}{ShuffleNetv2} & Fake & 0.9134 & 0.5722 \\
 & Real & 0.2343 & 0.9538 \\
 & Synthetic & 0.8782 & 0.9869 \\ \hline
\multirow{4}{*}{InceptionNetv3} & Fake & 0.6782 & 0.8912 \\
 & Real & 0.8708 & 0.6982 \\
 & Synthetic & 0.6213 & 0.9958 \\ \hline
\multirow{4}{*}{ViT Patch-16} & Fake & 0.9316 & 0.9956 \\
 & Real & 0.9903 & 0.9656 \\
 & Synthetic & 0.9993 & 0.9994 \\ \hline
\multirow{4}{*}{Efficientnet-b0} & Fake & 0.9422 & 0.9754 \\
 & Real & 0.9502 & 0.9706 \\
 & Synthetic & 0.9981 & 0.9993 \\ \hline
\multirow{4}{*}{MobileNetv2} & Fake & 0.7808 & 0.9894 \\
 & Real & 0.9783 & 0.8891 \\
 & Synthetic & 0.9962 & 0.9991 \\ \hline
\multirow{4}{*}{ResNet50} & Fake & 0.7078 & 0.6050 \\
 & Real & 0.3161 & 0.9174 \\
 & Synthetic & 0.6956 & 0.8373 \\ \hline
\multirow{4}{*}{DenseNet121} & Fake & 0.0334 & 0.9798 \\
 & Real & 0.1465 & 0.9625 \\
 & Synthetic & 0.9907 & 0.1431 \\ \hline
\multirow{4}{*}{VGG16} & Fake & 0.6017 & 0.5979 \\
 & Real & 0.2763 & 0.9194 \\
 & Synthetic & 0.6580 & 0.7507 \\ \hline
\end{tabular}
\label{tab:1}
\end{table}

\begin{table}[h]
\caption{Precision and accuracy metrics for each class across all trained models}
\label{tab:precision-accuracy}
\centering
\begin{tabular}{|l|l|c|c|}
\hline
\textbf{Model} & \textbf{Class} & \textbf{Precision} & \textbf{Accuracy} \\ \hline
\multirow{4}{*}{ShuffleNetv2} & Fake & 0.5163 & 0.6859 \\
 & Real & 0.7172 & 0.7140 \\
 & Synthetic & 0.9711 & 0.9507 \\ \hline
\multirow{4}{*}{InceptionNetv3} & Fake & 0.7571 & 0.8202 \\
 & Real & 0.5906 & 0.7557 \\
 & Synthetic & 0.9865 & 0.8709 \\ \hline
\multirow{4}{*}{ViT Patch-16} & Fake & 0.9905 & 0.9742 \\
 & Real & 0.9350 & 0.9738 \\
 & Synthetic & 0.9989 & 0.9994 \\ \hline
\multirow{4}{*}{EfficientNet-b0} & Fake & 0.9503 & 0.9643 \\
 & Real & 0.9417 & 0.9638 \\
 & Synthetic & 0.9986 & 0.9989 \\ \hline
\multirow{4}{*}{MobileNetv2} & Fake & 0.9736 & 0.9199 \\
 & Real & 0.8152 & 0.9188 \\
 & Synthetic & 0.9983 & 0.9982 \\ \hline
\multirow{4}{*}{ResNet50} & Fake & 0.4726 & 0.6393 \\
 & Real & 0.6569 & 0.7170 \\
 & Synthetic & 0.6813 & 0.7901 \\ \hline
\multirow{4}{*}{DenseNet121} & Fake & 0.4526 & 0.6643 \\
 & Real & 0.6611 & 0.6905 \\
 & Synthetic & 0.3663 & 0.4256 \\ \hline
\multirow{4}{*}{VGG16} & Fake & 0.4280 & 0.5992 \\
 & Real & 0.6315 & 0.7050 \\
 & Synthetic & 0.5689 & 0.7198 \\ \hline
\end{tabular}
\label{tab:2}
\end{table}

\begin{table}[htbp]
  \centering
  \caption{Class-averaged sensitivity, specificity, precision, and accuracy metrics for each model}
  \begin{tabular}{lcccc}
    \toprule
    \textbf{Model} & \textbf{Sensitivity} & \textbf{Specificity} & \textbf{Precision} & \textbf{Accuracy} \\
    \midrule
    InceptionNetv3 & 0.7234 & 0.8617 & 0.7781 & 0.8156 \\
    ViT Patch-16 & 0.9737 & 0.9869 & 0.9748 & 0.9825 \\
    MobileNetv2 & 0.9184 & 0.9592 & 0.9290 & 0.9456 \\
    VGG16 & 0.5120 & 0.7560 & 0.5428 & 0.6747 \\
    ShuffleNetv2 & 0.6753 & 0.8377 & 0.7349 & 0.7835 \\
    DenseNet121 & 0.3902 & 0.6951 & 0.4933 & 0.5935 \\
    ResNet50 & 0.5732 & 0.7866 & 0.6036 & 0.7154 \\
    EfficientNet-b0 & 0.9635 & 0.9818 & 0.9635 & 0.9757 \\
    \bottomrule
  \end{tabular}
  \label{tab:3}
\end{table}

\subsection{Image Property Analysis}

\begin{figure}[!t]
  \centering
  \includegraphics[scale=.17]{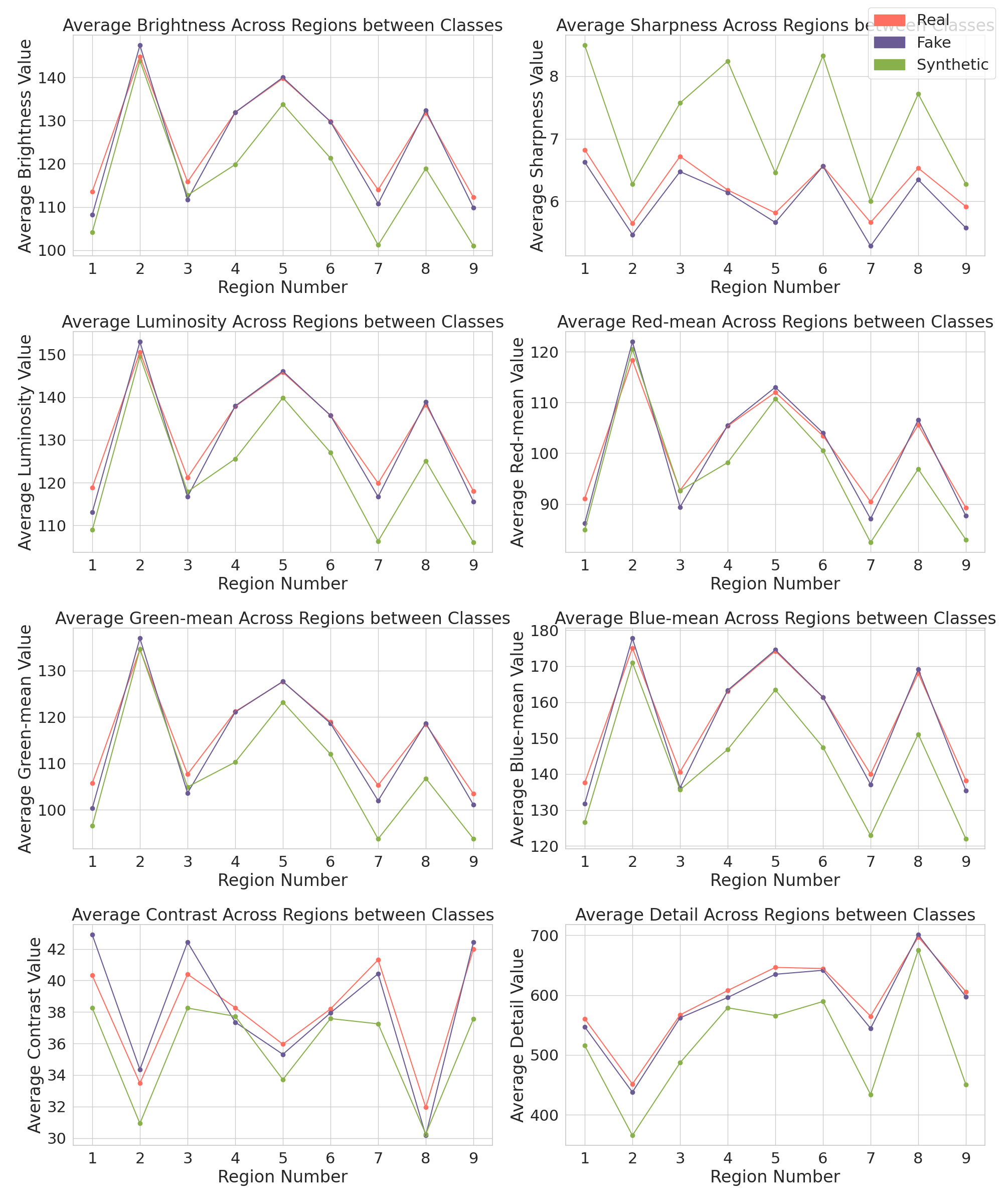}
  \caption{Line plots showcasing variation of average image property values for each class across different facial regions}
  \label{fig:6}
\end{figure}

As outlined in the line plots showcased in Fig. \ref{fig:6}, the average values of various image properties across all regions for each class of images consistently increase and decrease. However, it is also worth noting that the line plots corresponding to the synthetic class is typically below the curves of the other two classes, thus indicating lower average values across all image regions for each of the image properties, with the only exception being with regards to image sharpness whereby the line plot corresponding to the synthetic class is above the other curves. Despite this exception, the line plot corresponding to image sharpness still tends to stick to the overall trend in terms of how the values change as between image regions for each class. \par

Further insights can still be extracted from these line plots; for example, from region 1 to region 3, the average brightness, luminosity, red-mean, and blue mean values are approximately the same across all classes. Furthermore, we note that the biggest difference in average image property values between the synthetic class and the other two classes occurs within image regions 7, 8 and 9, apart from image contrast. Amongst middle image region ranges, the average image property values are almost identical for both the fake and real classes, as indicated within image regions 4, 5, 6, 7, 8. Finally, the line plots corresponding to image contrast property exhibits an overall similarity between average values across all classes, which points towards the similarity in image contrast levels between real, deepfake, and synthetic faces. \par

\begin{figure}[!t]
  \centering
  \includegraphics[scale=.17]{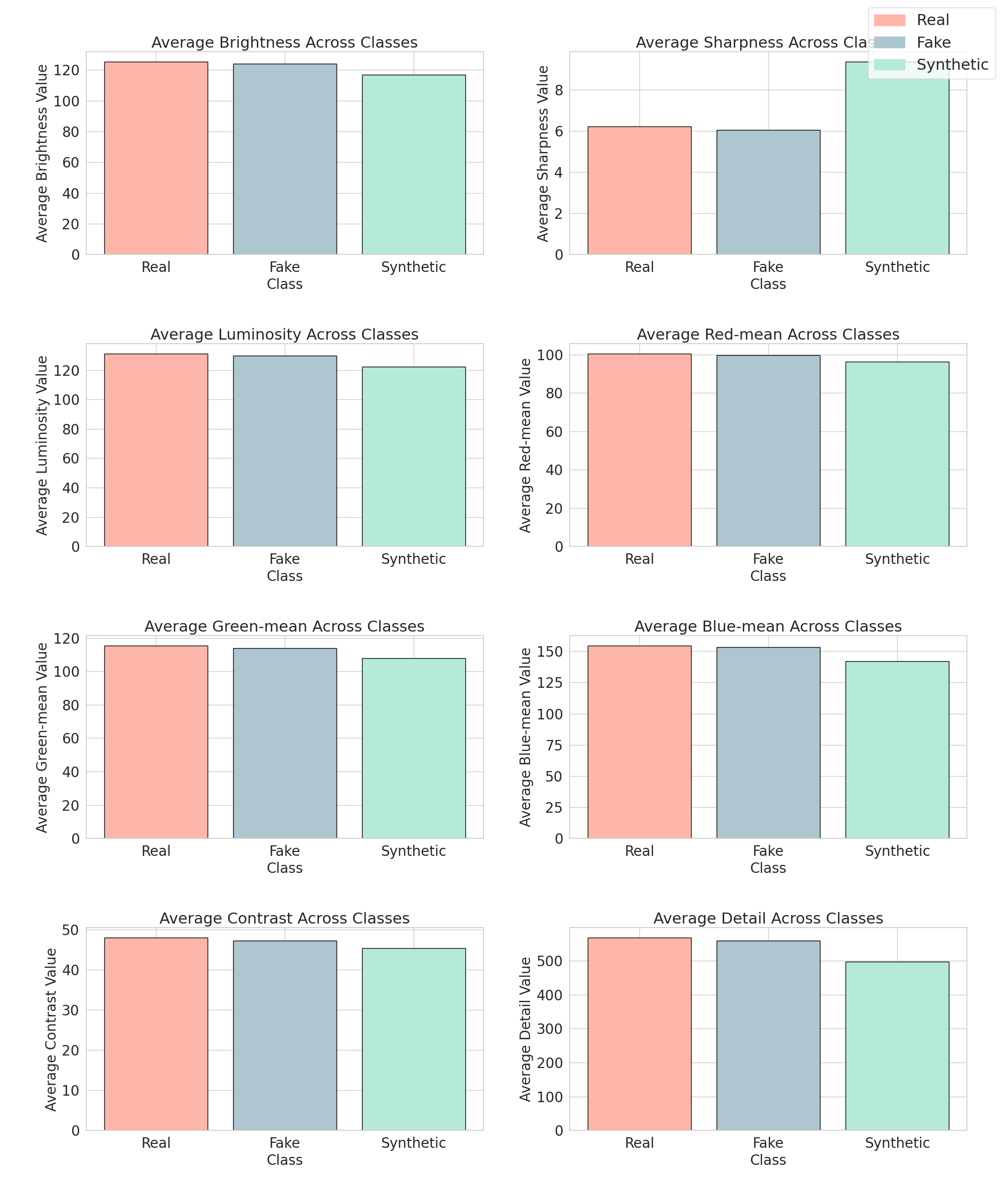}
  \caption{Average image property values for the tree classes: real, deepfake and synthetic face images}
  \label{fig:8}
\end{figure}

We can deduce that the greater differences in average image property values across image regions between the synthetic class and the other two classes should enable the deep learning models to detect the synthetic facial images with a higher level of accuracy. This hypothesis was also supported from our analysis of the deep learning models’ classification results. Consequently, we can deduce that six of the eight image properties (brightness, sharpness, luminosity, green-mean, blue-mean, and detail) seem to ease the synthetic image detection problem for the deep learning models as opposed to real and deepfake detection problems; these claims are supported by the results outlined in Fig. \ref{fig:6} and Fig. \ref{fig:8}. \par

Therefore, it is quite evident from the discussed line plots that facial images exhibit certain trends in visual properties. Each class generally follows the same trend: however, line plots corresponding to the synthetic class clearly stood out. This could be attributable to synthetic generative models attempting to be too perfect by not standing out in any part of the face and under compensates to try and give it a natural look. While such generative models do get close to creating realistic-looking images, the properties of such generated images remain notably different from deepfake and real images, e.g. the higher sharpness across the board. \par


\begin{figure}[!t]
  \centering
  \includegraphics[scale=.17]{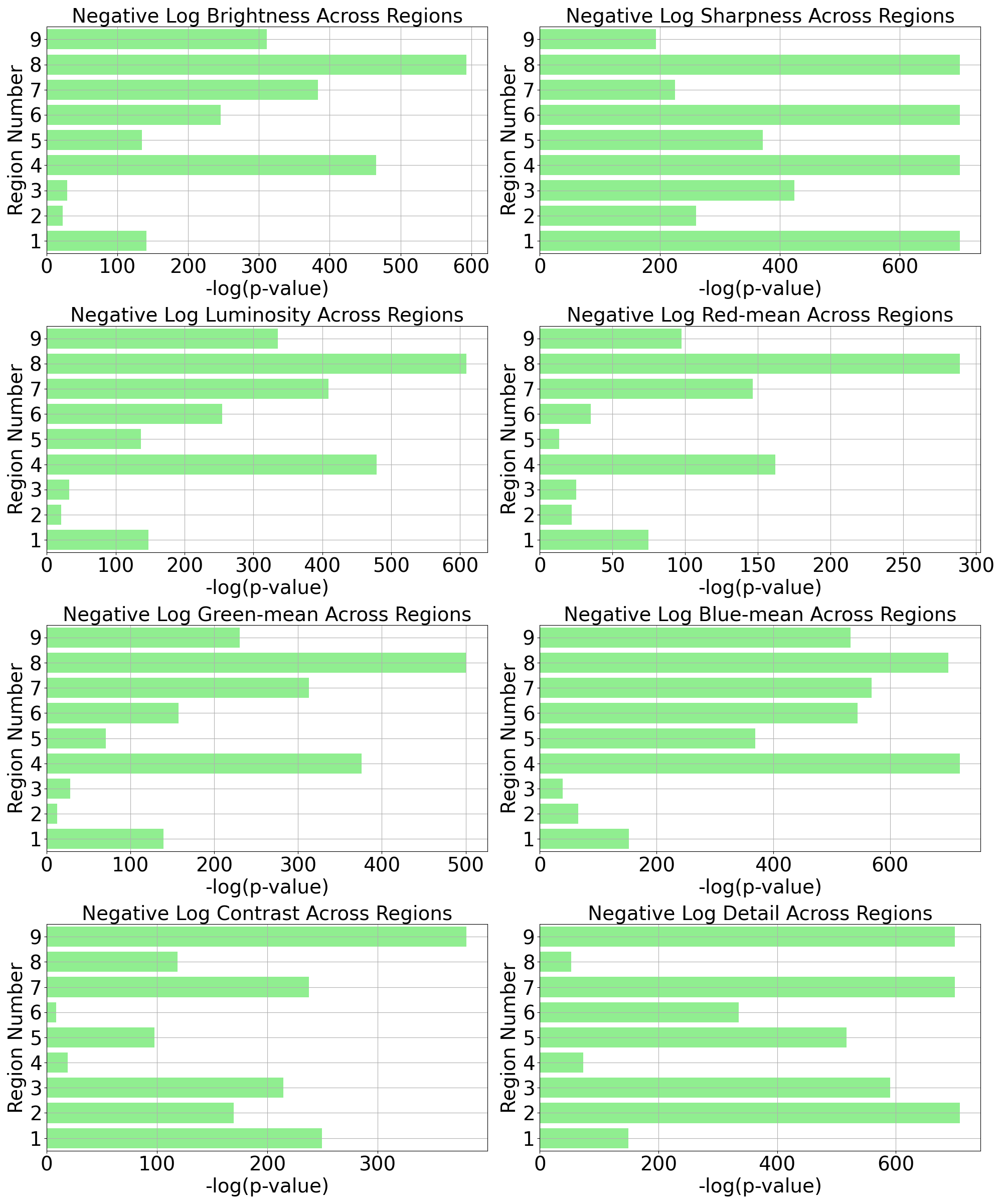}
  \caption{Negative log of p-values for various image properties when comparing different facial regions of real, deepfake, and synthetic faces.}
  \label{fig:7}
\end{figure}

Next when we look at the p-value plots in Fig. \ref{fig:7}. It is evident that no p-value is greater than 0.05. This means that the p-value for each property across regions is showing significant difference amongst the three image classes. We understand that the mean of no property within any region is the same and that this variability is unlikely due to chance. The p-values are so small that we had to use a negative logarithm scale to depict the differences between the various regions. It is also necessary to note here that some values were so small that their negative logarithm values were infinite and hence we capped them to a value. 


Observing the average image sharpness values, we note that the p-values are extremely low for four of the image regions, thus corresponding to high negative logarithmic values. This aligns with our expectations since the percentage difference between the different classes for this property was quite large as shown in Fig. \ref{fig:6}. A similar phenomenon occurs for the image detail property too. We also note that regions 1, 2 and 3 exhibit the highest p-values across all properties, and are therefore the least significant, except for image sharpness. This is further backed up by Fig. \ref{fig:6} where we see that in these regions, the values of the properties across all classes are similar. We also see that region 8 consistently has the lowest p-values except for in image contrast and detail, although this is not as evident from Fig. \ref{fig:6} compared to Fig. \ref{fig:7}. Another notable pattern is that the p-values mostly decrease from regions 4 to 8, except in contrast, detail and sharpness, while it also tends towards lower values in region 9 across all image properties. We can therefore derive from this that the top third of the face has fewer significant differences than the bottom two-thirds, in comparison; despite having small p-values. \par

Shifting our focus to analysis of the image as a whole instead of dealing with image regions, we note that the synthetic class still has lower average image property values compared to the other classes, in general, as shown in Fig. \ref{fig:8}. As was the case when analyzing based on image regions, sharpness remains an anomaly compared to the other image properties, and the percentage difference between the average image property values of the synthetic class and the other two classes remains notable. Furthermore, the real and deepfake classes still have very similar values across all image properties; the similarity between deepfake and real images can be attributed to the fact that deepfake images are generated using features of real images. The observed patterns back up the previously conducted analysis per image region. \par

\begin{figure}[!t]
  \centering
  \includegraphics[scale=.26]{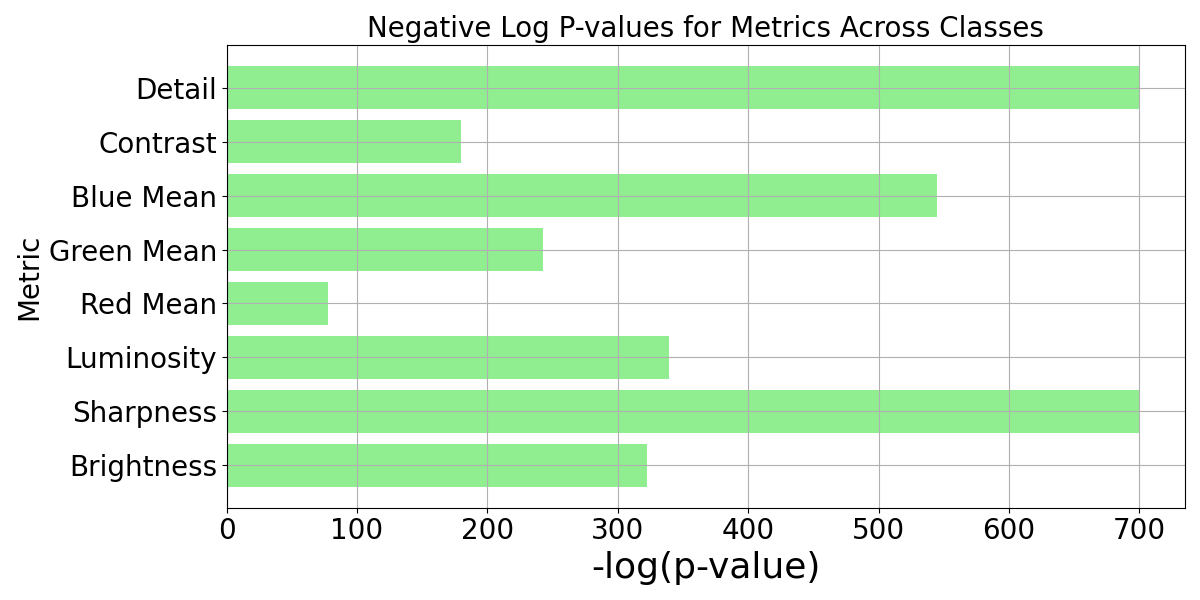}
  \caption{Negative log of p-values for various image properties when comparing real, deepfake, and synthetic faces.}
  \label{fig:9}
\end{figure}

The p-values are again the lowest for detail and sharpness as they were previously when analyzing based on image regions. In Fig. \ref{fig:9}, we can see that the highest p-value is exhibited by the red-mean image property. This is in line with the results in Fig. \ref{fig:7} as the negative logarithm of the p-values are lower for most of the regions for this property. The same pattern can be observed with the blue-mean value. The p-values of the various metrics across the classes when dealing with entire images are essentially the mean values across the image regions. Henceforth, we can safely say that these results also back up the ones concluded previously. \par

\section{Conclusion}
As inferred from our findings, we deduce that the trained InceptionNet-v3, ShuffleNet-v2, EfficientNet-b0, ResNet-50, ViT Patch-16, DenseNet-121, VGG-16, and MobileNetV2 deep learning models found it easier to distinguish synthetic images compared to deepfake and real images. We can attribute this to the clear difference in average image property values across the various features of synthetic facial images compared to deepfakes and real images, as deduced from the findings of our image property analysis. Despite the difference in average image property values, synthetic images still contain a strong element of realness, as indicated by the fact that the line plots corresponding to the average image property values for each property tended to follow the same trends as their real and deepfake counterparts. From our image property analysis, we also concluded that there was a resounding similarity in average image property values across the different image regions between deepfake and real facial images; this, coupled with the deep learning models’ difficulty in distinguishing between the two, highlights the very high level of verisimilitude that current deepfakes can attain. However, it is worth noting that, we were able to achieve high overall accuracy for the three image classes with some models such as ViT Patch-16, showing that these indeed are three different classes which can be treated separately from one another. With further developments in generative AI, deepfakes and synthetic images may reach a point where even AI struggles in distinguishing them from real images, which necessitates continuous future work on further improving the performance of deepfake and synthetic image detection models and algorithms. \par

\section*{Acknowledgments}
This work is supported in part by the grant FRG21-M-E94.

{\small
\bibliographystyle{IEEEtran}}
\bibliography{ReferencesFile}

\end{document}